\documentclass[sigconf]{acmart}
\usepackage[utf8]{inputenc}
\usepackage{subfigure}
\usepackage{multirow}
\usepackage{color, verbatim,caption}
\usepackage{hyperref}
\usepackage{enumitem}

\newtheorem{observation}[theorem]{Observation}
\usepackage{algorithm}
\usepackage{algorithmicx}
\usepackage{algpseudocode}
\usepackage{amsmath}
\usepackage{color}

\usepackage{array}
\newcolumntype{L}[1]{>{\raggedright\let\newline\\\arraybackslash\hspace{0pt}}m{#1}}
\newcolumntype{C}[1]{>{\centering\let\newline\\\arraybackslash\hspace{0pt}}m{#1}}
\newcolumntype{R}[1]{>{\raggedleft\let\newline\\\arraybackslash\hspace{0pt}}m{#1}}

\begin{document}


\title{Automatic Synonym Discovery with Knowledge Bases}

\author{Meng Qu}
\affiliation{%
  \institution{University of Illinois \\ at Urbana-Champaign}
}
\email{mengqu2@illinois.edu}

\author{Xiang Ren}
\affiliation{%
  \institution{University of Illinois \\ at Urbana-Champaign}
}
\email{xren7@illinois.edu}

\author{Jiawei Han}
\affiliation{%
  \institution{University of Illinois \\ at Urbana-Champaign}
}
\email{hanj@illinois.edu}

\begin{abstract}
Recognizing entity synonyms from text has become a crucial task in many entity-leveraging applications. However, discovering entity synonyms from domain-specific text corpora (\textit{e.g.}, news articles, scientific papers) is rather challenging. Current systems take an entity name string as input to find out other names that are synonymous, ignoring the fact that often times a name string can refer to multiple entities (\textit{e.g.}, ``apple'' could refer to both \textit{Apple Inc} and the fruit \textit{apple}). Moreover, most existing methods require training data manually created by domain experts to construct supervised-learning systems. 
In this paper, we study the problem of automatic synonym discovery with knowledge bases, that is, identifying synonyms for \textit{knowledge base entities} in a given domain-specific corpus.
The manually-curated synonyms for each entity stored in a knowledge base not only form a set of name strings to \textit{disambiguate} the meaning for each other, but also can serve as ``\textit{distant}'' supervision to help determine important features for the task.
We propose a novel framework, called \textbf{DPE}, to integrate two kinds of mutually-complementing signals for synonym discovery, \textit{i.e.}, \textit{distributional features} based on corpus-level statistics and \textit{textual patterns} based on local contexts. In particular, DPE jointly optimizes the two kinds of signals in conjunction with distant supervision, so that they can mutually enhance each other in the training stage. At the inference stage, both signals will be utilized to discover synonyms for the given entities. Experimental results prove the effectiveness of the proposed framework.

\end{abstract}



\copyrightyear{2017} 
\acmYear{2017} 
\setcopyright{acmcopyright}
\acmConference{KDD'17}{}{August 13-17, 2017, Halifax, NS, Canada}
\acmPrice{15.00}
\acmDOI{10.1145/3097983.3098185}
\acmISBN{978-1-4503-4887-4/17/08}

\fancyhead{}

\maketitle

\section{Introduction}
\label{sec::intro}

People often have a variety of ways to refer to the same real-world entity, forming different \textit{synonyms} for the entity (\textit{e.g.}, entity \textit{United States} can be referred using ``\textit{America}'' and ``\textit{USA}'').
Automatic synonym discovery is an important task in text analysis and understanding, as the extracted synonyms (\textit{i.e.} the alternative ways to refer to the same entity) can benefit many downstream applications~\cite{zeng2012synonym,angheluta2002use,wu2010open,xie2015incorporating}. 
For example, by forcing synonyms of an entity to be assigned in the same topic category, one can constrain the topic modeling process and yield topic representations with higher quality~\cite{xie2015incorporating}. 
Another example is in document retrieval~\cite{voorhees1994query}, where we can leverage entity synonyms to enhance the process of query expansion, and thus improve the retrieval performances.

One straightforward approach for obtaining entity synonyms is to leverage publicly available knowledge bases such as Freebase and WordNet, in which popular synonyms for the entities are manually curated by human crowds.
However, the coverage of knowledge bases can be rather limited, especially on some newly emerging entities, as the manual curation process entails high costs and is not scalable. 
For example, the entities in Freebase have only 1.1 synonyms on average.
To increase the synonym coverage, we expect to automatically extract more synonyms that are not in knowledge bases from massive, domain-specific text corpora.
Many approaches address this problem through supervised~\cite{weeds2014learning,roller2014inclusive,wang2015solving} or weakly supervised learning~\cite{snow2004learning,nakashole2012patty}, which treat some manually labeled synonyms as seeds to train a synonym classifier or detect some local patterns for synonym discovery.
Though quite effective in practice, such approaches still rely on careful seed selections by humans.

To retrieve training seeds automatically, recently there is a growing interest in the distant supervision strategy, which aims to automatically collect training seeds from existing knowledge bases.
The typical workflow is: i) detect entity mentions from the given corpus, ii) map the detected entity mentions to the entities in a given knowledge base, iii) collect training seeds from the knowledge base.
Such techniques have been proved effective in a variety of applications, such as relation extraction~\cite{mintz2009distant}, entity typing~\cite{ren2015clustype} and emotion classification~\cite{purver2012experimenting}.
Inspired by such strategy, a promising direction for automatic synonym discovery could be collecting training seeds (\textit{i.e.}, a set of synonymous strings) from knowledge bases.

\begin{figure}
    \centering
    \includegraphics[width=0.45\textwidth]{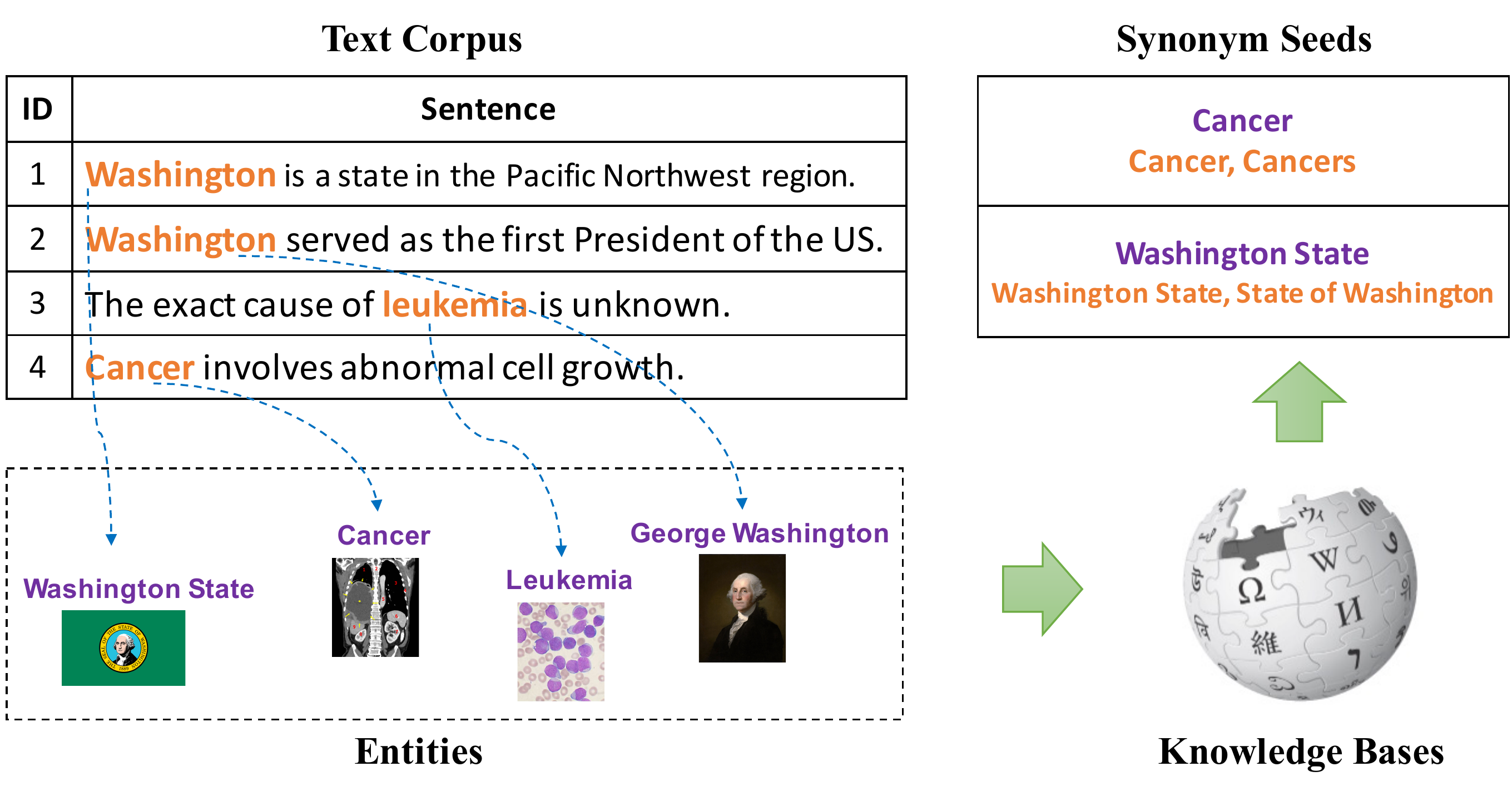}
    \caption{Distant supervision for synonym discovery. We link entity mentions in text corpus to knowledge base entities, and collect training seeds from knowledge bases.}
    \label{fig::example}
    \vspace{-0.3cm}
\end{figure}

Although distant supervision helps collect training seeds automatically, it also poses a challenge due to the string ambiguity problem, that is, the same entity surface strings can be mapped to different entities in knowledge bases. For example, consider the string ``\textit{Washington}'' in Figure~\ref{fig::example}. The ``\textit{Washington}'' in the first sentence represents a state of the United States; while in the second sentence it refers to a person.
As some strings like ``\textit{Washington}'' have ambiguous meanings, directly inferring synonyms for such strings may lead to a set of synonyms for multiple entities. For example, the synonyms of entity \textit{Washington} returned by current systems may contain both the state names and person names, which is not desirable. To address the challenge, instead of using ambiguous strings as queries, a better way is using some specific concepts as queries to disambiguate, such as entities in knowledge bases.

This motivated us to define a new task: \emph{automatic synonym discovery for entities with knowledge bases}. Given a domain-specific corpus, we aim to collect existing name strings of entities from knowledge bases as seeds. For each query entity, the existing name strings of that entity can disambiguate the meaning for each other, and we will let them vote to decide whether a given candidate string is a synonym of the query entity.
Based on that, the key task for this problem is to predict whether a pair of strings are synonymous or not.
For this task, the collected seeds can serve as supervision to help determine the important features.
However, as the synonym seeds from knowledge bases are usually quite limited, how to use them effectively becomes a major challenge. There are broadly two kinds of efforts towards exploiting a limited number of seed examples.

The distributional based approaches~\cite{weeds2014learning,roller2014inclusive,mikolov2013distributed,pennington2014glove,wang2015solving} consider the corpus-level statistics, and they assume strings which often appear in similar contexts are likely to be synonyms. For example, the strings ``\textit{USA}'' and ``\textit{United States}'' are usually mentioned in similar contexts, and they are the synonyms of the country \textit{USA}. Based on the assumption, the distributional based approaches usually represent strings with their distributional features, and treat the synonym seeds as labels to train a classifier, which predicts whether a given pair of strings are synonymous or not. Since most synonymous strings will appear in similar contexts, such approaches usually have high recall. However, such strategy also brings some noise, since some non-synonymous strings may also share similar contexts, such as ``\textit{USA}'' and ``\textit{Canada}'', which could be labeled as synonyms incorrectly.

Alternatively, the pattern based approaches~\cite{snow2004learning,sun2010semi,qian2009semi,hearst1992automatic} consider the local contexts, and they infer the relation of two strings by analyzing sentences mentioning both of them. For example, from the sentence \emph{``The United States of America is commonly referred to as America.''}, we can infer that ``\textit{United States of America}'' and ``\textit{America}'' have the synonym relation; while the sentence \emph{``The USA is adjacent to Canada''} may imply that ``\textit{USA}'' and ``\textit{Canada}'' are not synonymous. To leverage this observation, the pattern based approaches will extract some textual patterns from sentences in which two synonymous strings co-occur, and discover more synonyms with the learned patterns.
Different from the distributional based approaches, the pattern based approaches can treat the patterns as concrete evidences to support the discovered synonyms, which are more convincing and interpretable. However, as many synonymous strings will not be co-mentioned in any sentences, such approaches usually suffer from low recall.

Ideally, we would wish to combine the merits of both approaches, and in this paper we propose such a solution named DPE (distributional and pattern integrated embedding framework). Our framework consists of a distributional module and a pattern module. The distributional module predicts the synonym relation from the \emph{global} distributional features of strings; while in the pattern module, we aim to discover synonyms from the \emph{local} contexts. 
Both modules are built on top of some string embeddings, which preserve the semantic meanings of strings. During training, both modules will treat the embeddings as features for synonym prediction, and in turn update the embeddings based on the supervision from synonym seeds. The string embeddings are shared across the modules, and therefore each module can leverage the knowledge discovered by the other module to improve the learning process.

To discover missing synonyms for an entity, one may directly rank all candidate strings with both modules. However, such strategy can have high time costs, as the pattern module needs to extract and analyze all sentences mentioning a pair of given strings when predicting their relation. To speed up synonym discoveries, our framework will first utilize the distributional module to rank all candidate strings, and extract a set of top ranked candidates as high-potential ones. After that, we will re-rank the high-potential candidates with both modules, and treat the top ranked candidates as the discovered synonyms.

\begin{figure}
    \centering
    \includegraphics[width=0.45\textwidth]{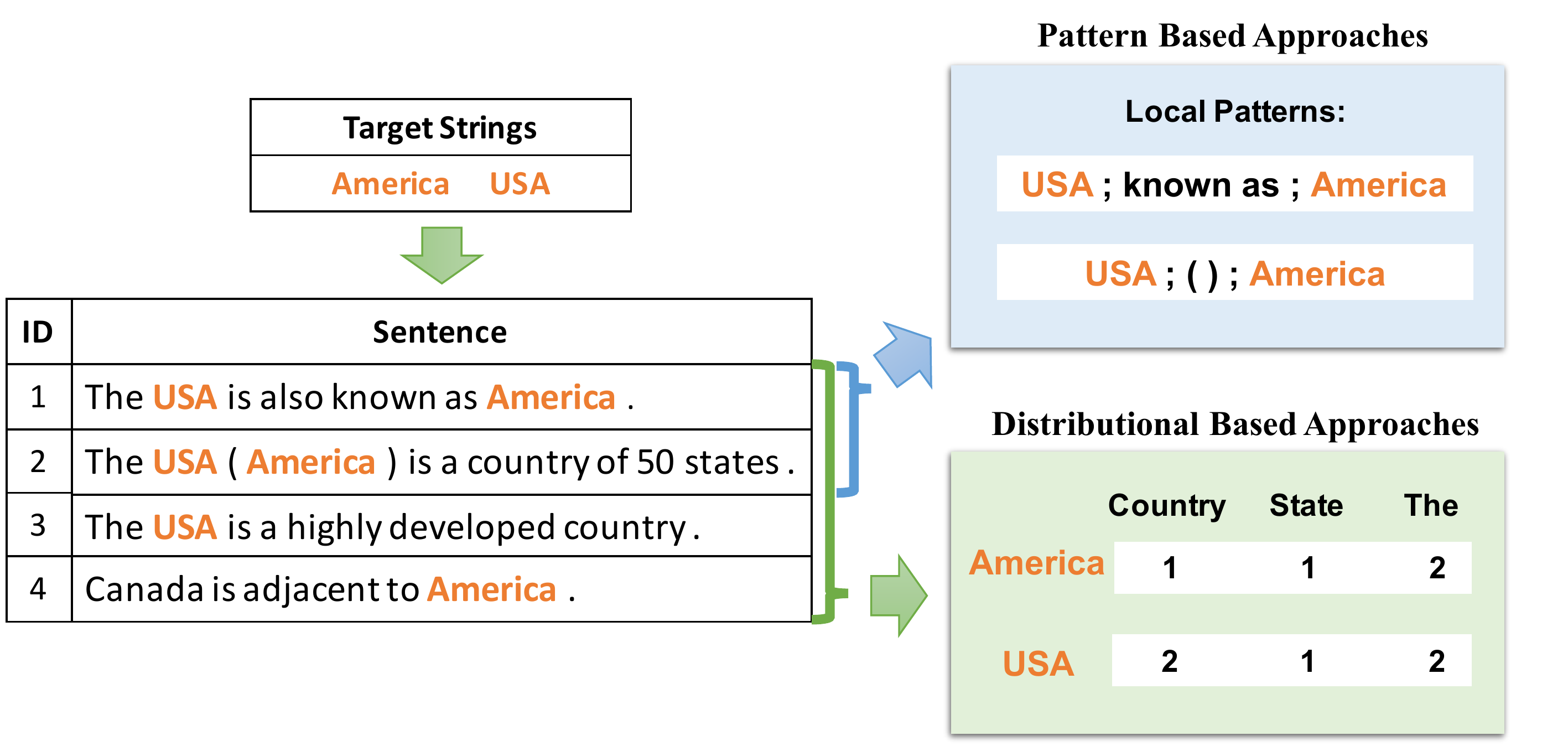}
    \caption{
    Comparison of the distributional based and pattern based approaches. To predict the relation of two target strings, the distributional based approaches will analyze their distributional features, while the pattern based approaches will analyze the local patterns extracted from sentences mentioning both of the target strings.
    }
    \label{fig::comparison}
    \vspace{-0.3cm}
\end{figure}

The major contributions of the paper are summarized as follows:
\begin{itemize}[leftmargin=*,noitemsep,nolistsep]
	\item We propose to study the problem of \emph{automatic synonym discovery with knowledge bases}, \textit{i.e.}, aiming to discover missing synonyms for entities by collecting training seeds from knowledge bases.
	\item We propose a novel approach DPE, which naturally integrates the distributional based approaches and the pattern based approaches for synonym discovery.
	\item We conduct extensive experiments on the real world text corpora. Experimental results prove the effectiveness of our proposed approach over many competitive baseline approaches. 
\end{itemize}

\begin{figure*}
    \centering
    \includegraphics[width=0.9\textwidth]{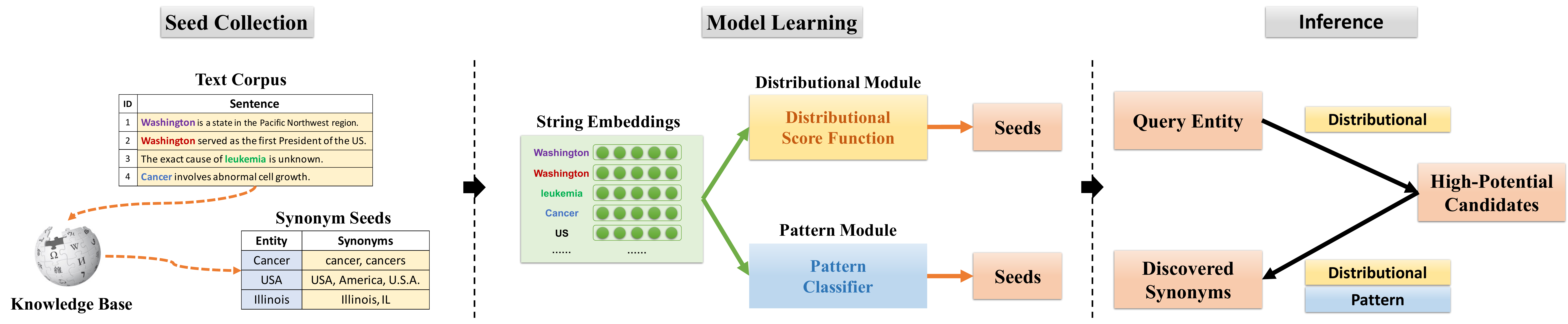}
    \caption{Framework Overview.}
    \label{fig::framework}
\end{figure*}

\section{Problem Definition}
\label{sec::definition}

In this section, we define several concepts and our problem:

\smallskip
\noindent \textsf{\textbf{Synonym.}}
A synonym is a string (\textit{i.e.}, word or phrase) that means exactly or nearly the same as another string in the same language~\cite{stanojevic2009cognitive}. Synonyms widely exist in human languages. For example, ``\textit{Aspirin}'' and ``\textit{Acetylsalicylic Acid}'' refer to the same drug; ``\textit{United States}'' and ``\textit{USA}'' represent the same country. All these pairs of strings are synonymous.

\smallskip
\noindent \textsf{\textbf{Entity Synonym.}}
For an entity, its synonym refers to strings that can be used as alternative names to describe that entity.
For example, both the strings ``\textit{USA}'' and ``\textit{United States}'' serve as alternative names of the entity \textit{United States}, and therefore they are the synonyms of this entity.

\smallskip
\noindent \textsf{\textbf{Knowledge Base.}}
A knowledge base consists of some manually constructed facts about a set of entities. In this paper, we only focus on the existing entity synonyms provided by knowledge bases, and we will collect those existing synonyms as training seeds to help discover other missing synonyms.

\smallskip
\noindent \textsf{\textbf{Problem Definition.}}

Given the above concepts, we formally define our problem as follows.

\begin{definition}
	\label{def::prob}
	\textbf{(Problem Definition)}
	\textsl{Given a knowledge base $\mathcal{K}$ and a text corpus $\mathcal{D}$, our problem aims to extract the missing synonyms for the query entities.
	}
\end{definition}

\section{Framework}

In this section, we introduce our approach DPE for entity synonym discovery with knowledge bases. To infer the synonyms of a query entity, we leverage its name strings collected from knowledge bases to disambiguate the meaning for each other, and let them vote to decide whether a given candidate string is a synonym of the query entity. Therefore, the key task for this problem is to predict whether a pair of strings are synonymous or not. For this task, the synonym seeds collected from knowledge bases can serve as supervision to guide the learning process. However, as the number of synonym seeds is usually small, how to leverage them effectively is quite challenging.
Existing approaches either train a synonym classifier with the distributional features, or learn some textual patterns for synonym discovery, which cannot exploit the seeds sufficiently.

To address this challenge, our framework naturally integrates the distributional based approaches and the pattern based approaches. Specifically, our framework consists of a distributional module and a pattern module. Given a pair of target strings, the distributional module predicts the synonym relation from the global distributional features of each string; while the pattern module considers the local contexts mentioning both target strings. During training, both modules will mutually enhance each other. At the inference stage, we will leverage both modules to find high-quality synonyms for the query entities.

\smallskip
\noindent \textsf{\textbf{Framework Overview.}}
The overall framework of DPE (Figure~\ref{fig::framework}) is summarized below:
\begin{enumerate}[leftmargin=*,noitemsep,nolistsep]
    \item Detect entity mentions in the given text corpus and link them to entities in the given knowledge base. Collect synonym seeds from knowledge bases as supervision.
    \item Jointly optimize the distributional and the pattern modules. The distributional module predicts synonym relations with the global distributional features, while the pattern module considers the local contexts mentioning both target strings.
    \item Discover missing synonyms for the query entities with both the distributional module and the pattern module.
\end{enumerate}

\subsection{Synonym Seed Collection}
To automatically collect synonym seeds, our approach will first detect entity mentions (strings that represent entities) in the given text corpus and link them to entities in the given knowledge base. After that, we will retrieve the existing synonyms in knowledge bases as our training seeds. An illustrative example is presented in Figure~\ref{fig::example}.

Specifically, we first apply existing named-entity recognition (NER) tools~\cite{manning-EtAl:2014:P14-5}\footnote{~\url{http://stanfordnlp.github.io/CoreNLP/}} to detect entity mentions and phrases in the given text corpus. Then some entity linking techniques such as the DBpedia Spotlight~\cite{isem2013daiber}\footnote{~\url{https://github.com/dbpedia-spotlight/dbpedia-spotlight}} are applied, which will map the detected entity mentions to the given knowledge base. During entity linking, some mentions can be linked to incorrect entities, leading to false synonym seeds. To remove such seeds, for each mention and its linked entity, if the surface string of that mention is not in the existing synonym list of that entity, we will remove the link between the mention and the entity,

After entity mention detection and linking, the synonym seeds will be collected from the linked corpus. Specifically, for each entity, we collect all mentions linked to that entity, and treat all corresponding surface strings as synonym seeds.

\subsection{Methodology Overview}

After extracting synonym seeds from knowledge bases, we formulate an optimization framework to jointly learn the distributional module and the pattern module.

To preserve the semantic meanings of different strings, our framework introduces a low-dimensional vector (\textit{a.k.a.} embedding) to represent each entity surface string (\textit{i.e.}, strings that are linked to entities in knowledge bases) and each unlinkable string (\textit{i.e.}, words and phrases that are not linked to any entities). For the same strings that linked to different entities, as they have different semantic meanings, we introduce different embeddings for them. For example, the string ``\textit{Washinton}'' can be linked to a state or a person, and we use two embeddings to represent \textit{Washinton} (state) and \textit{Washinton} (person) respectively.

The two modules of our framework are built on top of these string embeddings. Specifically, both modules treat the embeddings as features for synonym prediction, and in turn update the embeddings based on the supervision from the synonym seeds, which may bring stronger predictive abilities to the learned embeddings. Meanwhile, since the string embeddings are shared between the two modules, each module is able to leverage the knowledge discovered by the other module, so that the two modules can mutually enhance to improve the learning process.

The overall objective of our framework is summarized as follows:
\begin{equation}
	\label{eqn::obj-all}
	O=O_{D} + O_{P},
\end{equation}
where $O_{D}$ is the objective of the distributional module and $O_{P}$ is the objective of the pattern module. Next, we introduce the details of each module.

\subsubsection{\textbf{Distributional Module}}
The distributional module of our framework considers the global distributional features for synonym discovery. The module consists of an unsupervised part and a supervised part. In the unsupervised part, a co-occurrence network encoding the distributional information of strings will be constructed, and we try to preserve the distributional information into the string embeddings. Meanwhile in the supervised part, the synonym seeds will be used to learn a distributional score function, which takes string embeddings as features to predict whether two strings are synonymous or not.

\smallskip
\noindent \textsf{\textbf{Unsupervised Part.}}
In the unsupervised part, we first construct a co-occurrence network between different strings, which captures their distributional information. Formally, all strings (\textit{i.e.}, entity surface strings and other unlinkable strings) within a sliding window of a certain size $w$ in the text corpus are considered to be co-occurring with each other. The weight for each pair of strings in the co-occurrence network is defined as their co-occurrence count.

After network construction, we aim to preserve the encoded distributional information into the string embeddings, so that strings with similar semantic meanings will have similar embeddings. To preserve the distributional information, we observe that the co-occurrence counts of strings are related to the following factors.
\begin{observation}[Co-occurrence Observation]
\label{obs::co-occur}
(1) If two strings have similar semantic meanings, then they are more likely to co-occur with each other.
(2) If a string tends to appear in the context of another one, then they tend to co-occur frequently.
\end{observation}
The above observation is quite intuitive. If two strings have similar semantic meanings, they are more likely to be mentioned in the same topics, and therefore have a larger co-occurrence probability. For example, the strings ``\textit{data mining}'' and ``\textit{text mining}'' are highly correlated, while they have quite different meanings from the word ``\textit{physics}'', and we can observe that the co-occurrence chances between ``\textit{data mining}'' and ``\textit{text mining}'' are much larger than those between ``\textit{data mining}'' and ``\textit{physics}''.
On the other hand, some string pairs with very different meanings may also have large co-occurrence counts, when one tends to appear in the context of the other one. For example, the word ``\textit{capital}'' often appears in the context of ``\textit{USA}'', even they have very different meanings.

To exploit the above observation, for each string $u$, besides its embedding vector $\mathbf{x}_{u}$, we also introduce a context vector $\mathbf{c}_{u}$, which describes what kinds of strings are likely co-mentioned with $u$. Given a pair of strings $(u, v)$, we model the conditional probability $p(u|v)$ as follows:
\begin{equation}
	\label{eqn::softmax}
	p(u|v)=\frac{\exp(\mathbf{x}_u^T \mathbf{x}_v + \mathbf{x}_u^T \mathbf{c}_v)}{Z},
\end{equation}
where $Z$ is a normalization term. We see that if $u$ and $v$ have similar embedding vectors, meaning they have similar semantic meanings, the first part ($\mathbf{x}_u^T \mathbf{x}_v$) of the equation will be large, leading to a large conditional probability, which corresponds to the first observation~\ref{obs::co-occur}. On the other hand, if the embedding vector of $u$ is similar to the context vector of $v$, meaning $u$ tends to appear in the context of $v$, the second part ($\mathbf{x}_u^T \mathbf{c}_v$) becomes large, which also leads to a large conditional probability, and this process corresponds to the second observation~\ref{obs::co-occur}.

To preserve the distributional information of strings, we expect the estimated distribution $p(\cdot|v)$ to be close to the empirical distribution $p'(\cdot|v)$ (\textit{i.e.}, $p'(u|v)=w_{u,v}/d_v$, where $w_{u,v}$ is the co-occurrence count between $u$ and $v$, and $d_v$ is the degree of $v$ in the network) for each string $v$. Therefore, we minimize the KL distance between $p(\cdot|v)$ and $p'(\cdot|v)$, which is equivalent to the following objective~\cite{tang2015line}:
\begin{equation}
	\label{eqn::obj-occur}
	L_C=\sum_{u,v \in V}w_{u,v}\log p(u|v),
\end{equation}
where $V$ is the vocabulary of all strings.

Directly optimizing the above objective is computational expensive since it involves traversing all strings in the vocabulary when computing the conditional probability. Therefore, we leverage the negative sampling techniques~\cite{mikolov2013distributed} to speed up the learning process, which modify the conditional probability $p(u|v)$ in Eqn.~\ref{eqn::obj-occur} as follows:
\begin{equation}
\begin{footnotesize}
	\label{eqn::obj-occur-final}
	\log \sigma(\mathbf{x}_u^T \mathbf{x}_v + \mathbf{x}_u^T \mathbf{c}_v)+
	\sum_{n=1}^NE_{u_n\sim P_{neg}(u)}[1-\log \sigma(\mathbf{x}_{u_n}^T \mathbf{x}_v + \mathbf{x}_{u_n}^T \mathbf{c}_v)],
\end{footnotesize}
\end{equation}
where $\sigma(x)=1/(1+\exp(-x))$ is the sigmoid function. The first term tries to maximize the probabilities of some observed string pairs, while the second term tries to minimize the probabilities of $N$ noisy pairs, and $u_n$ is sampled from a noisy distribution $P_{neg}(u) \propto d_u^{3/4}$ and $d_u$ is the degree of string $u$ in the network.

\smallskip
\noindent \textsf{\textbf{Supervised Part.}}
The unsupervised part of the distributional module can effectively preserve the distributional information of strings into the learned string embeddings. In the supervised part, we will utilize the collected synonym seeds to train a distributional score function, which treats the string embeddings as features to predict whether two strings have the synonym relation or not.

To measure how likely two strings are synonymous, we introduce a score for each pair of strings. Inspired by the existing study~\cite{yang2014embedding}, we use the following bilinear function to define the score of a string pair $(u,v)$:
\begin{equation}
	\label{eqn::score_d}
	Score_D(u,v) = \mathbf{x}_u \mathbf{W}_D \mathbf{x}_v^T,
\end{equation}
where $\mathbf{x}_u$ is the embedding of string $u$, $\mathbf{W}_D$ is a parameter matrix for the score function. Due to the efficiency issue, in this paper we constrain $\mathbf{W}_D$ as a diagonal matrix.

To learn the parameters $\mathbf{W}_D$ in the score function, we expect that the synonymous string pairs could have larger scores than those randomly sampled pairs. Therefore we adopt the following ranking based objective for learning:
\begin{equation}
	\label{eqn::obj-synonym}
	L_S = \sum_{(u,v) \in S_{seed}} \sum_{v' \in V}\min(1,Score_D(u,v) - Score_D(u,v')),
\end{equation}
where $S_{seed}$ is the set of synonymous string pairs, $v'$ is a string randomly sampled from the string vocabulary. By maximizing the above objective, the learned parameter matrix $\mathbf{W}_D$ will be able to distinguish those synonymous pairs from others. Meanwhile, we will update the string embeddings to maximize the objective, which will bring more predictive abilities to the learned embeddings.

\subsubsection{\textbf{Pattern Module}}

\begin{figure}
    \centering
    \includegraphics[width=0.45\textwidth]{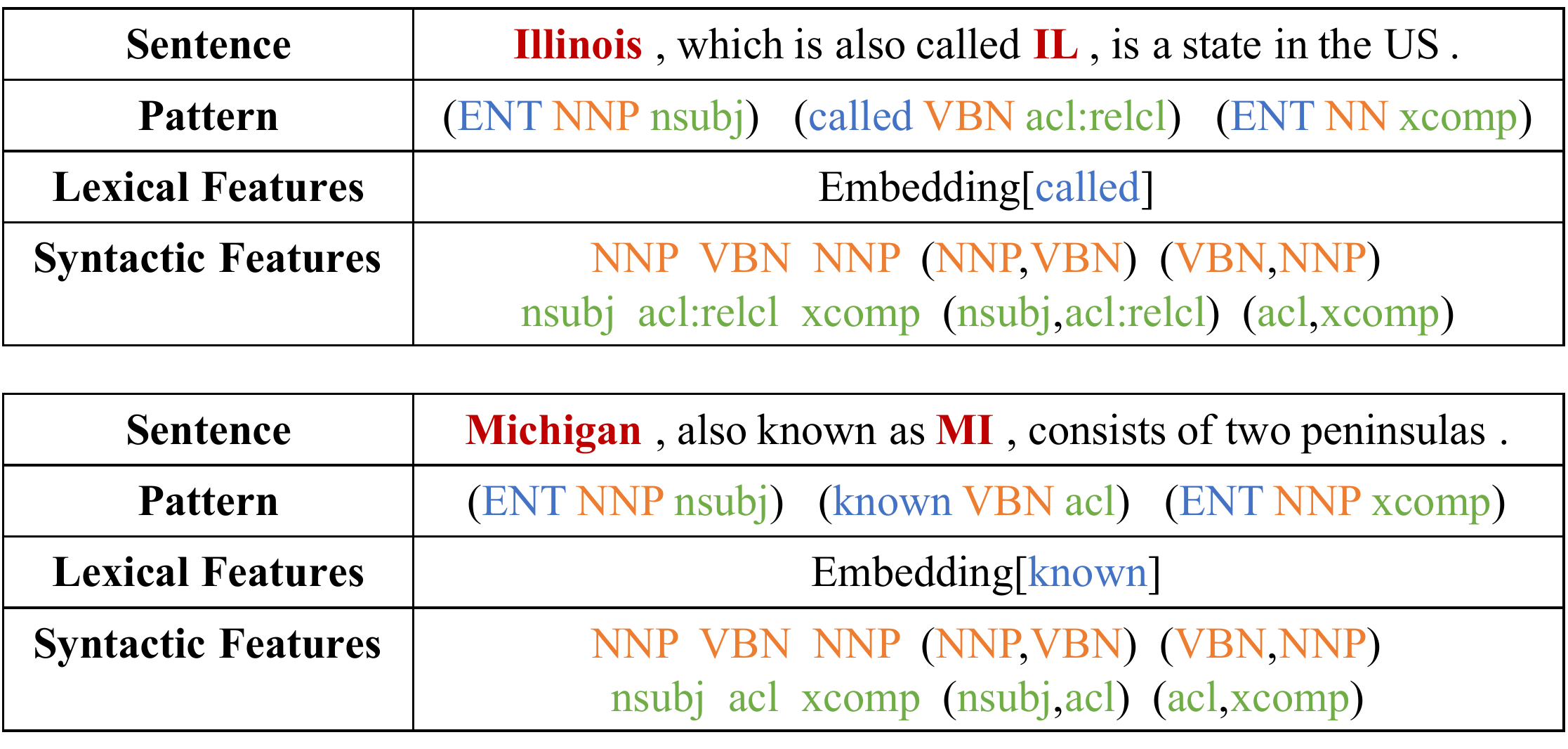}
    \caption{
    Examples of patterns and their features. For a pair of target strings (red ones) in each sentence, we define the pattern as the $<$token, POS tag, dependency label$>$ triples in the shortest dependency path. We collect both lexical features and syntactic features for pattern classification.
    }
    \label{fig::pattern}
    \vspace{-0.3cm}
\end{figure}

For a pair of target strings, the pattern module of our framework predicts their relation from the sentences mentioning both of them. We achieve this by extracting a pattern from each of such sentences, and collecting some lexical features and syntactic features to represent each pattern. Based on the extracted features, a pattern classifier is trained to predict whether a pattern expresses the synonym relation between the target strings. Finally, we will integrate all prediction results from these patterns to decide the relation of the target strings.

We first introduce the definition of the pattern used in our framework.
Following existing pattern based approaches~\cite{nakashole2012patty,yahya2014renoun}, given two target strings in a sentence, we define the pattern as the sequence of $<$lexical string, part-of-speech tag, dependency label$>$ triples collected from the shortest dependency path connecting the two strings. Two examples can be found in Figure~\ref{fig::pattern}.

For each pattern, we will extract some features and predict whether this pattern expresses the synonym relation. We expect that the extracted features could well capture the functional correlations between patterns. In other words, patterns expressing synonym relations should have similar features. For example, consider the two sentences in Figure~\ref{fig::pattern}. The patterns in both sentences express the synonym relation between the target strings (strings with the red color), and therefore we anticipate that the two patterns could have similar features.

Towards this goal, we extract both lexical and syntactic features for each pattern. For the lexical features, we average all embeddings of strings in a pattern as the features. As the string embeddings can well preserve the semantic meanings of strings, such lexical features can effectively capture the semantic correlations between different patterns. Take the sentences in Figure~\ref{fig::pattern} as an example. Since the strings ``\textit{called}'' and ``\textit{known}'' usually appear in similar contexts, they will have quite similar embeddings, and therefore the two patterns will have similar lexical features, which is desirable. For the syntactic features, we expect that they can capture the syntactic structures of the patterns. Therefore for each pattern, we treat all n-grams ($1 \leq n \leq N$) in the part-of-speech tag sequence and the dependency label sequence as its syntactic features. Some example are presented in Figure~\ref{fig::pattern}, where we set $N$ as 2.

Based on the extracted features, a pattern classifier will be trained, which predicts whether a pattern expresses the synonym relation. To collect positive examples for training, we extract patterns from all sentences mentioning a pair of synonymous strings, and treat these patterns as positive examples. For the negative examples, we randomly sample some string pairs without the synonym relation, and treat the corresponding patterns as negative ones. We select the linear logistic classifier for classification. Given a pattern $pat$ and its feature vector $\mathbf{f}_{pat}$, we define the probability that pattern $pat$ expresses the synonym relation as follows:
\begin{equation}
	\label{eqn::obj-synonym}
	P(y_{pat}=1|pat)=\frac{1}{1+\exp(-\mathbf{W_P}^T\mathbf{f}_{pat})},
\end{equation}
where $\mathbf{W_P}$ is the parameter vector of the classifier.
We learn $\mathbf{W_P}$ by maximizing the log-likelihood objective function, which is defined as below:
\begin{equation}
	\label{eqn::obj-synonym}
	O_{P}=\sum_{pat \in S_{pat}} \log P(y_{pat}|pat),
\end{equation}
where $S_{pat}$ is the set of all training patterns, $y_{pat}$ is the label of pattern $pat$. By maximizing the above objective, the learned classifier can effectively predict whether a pattern expresses the synonym relation or not. Meanwhile, we will also update the string embeddings during training, and therefore the learned string embeddings will have better predictive abilities for the synonym discovery problem.

After learning the pattern classifier, we can use it for synonym prediction. Specifically, for a pair of target strings $u$ and $v$, we first collect all sentences mentioning both strings, and extract corresponding patterns from them, then we measure the possibility that $u$ and $v$ are synonymous using the following score function $Score_P(u,v)$:
\begin{equation}
	\label{eqn::score_p}
	Score_P(u,v) = \frac{\sum_{pat \in S_{pat}(u,v)}P(y_{pat}=1|pat)}{|S_{pat}(u,v)|},
\end{equation}
where $S_{pat}(u,v)$ is the set of all corresponding patterns. Basically, our approach will classify all corresponding patterns, and different patterns will vote to decide whether $u$ and $v$ are synonymous.

\section{Model Learning and Inference}
In this section, we introduce our optimization algorithm and how we discover missing synonyms for entities.

\smallskip
\noindent \textsf{\textbf{Optimization Algorithm.}}
The overall objective function of our framework consists of three parts. Two of them ($L_C$ and $L_S$) are from the distributional module and the other one ($O_P$) is from the pattern module. To optimize the objective, we adopt the edge sampling strategy~\cite{tang2015line}. In each iteration, we alternatively sample a training example from the three parts, and then update the corresponding parameters. We summarize the optimization algorithm in Algorithm~\ref{algo::optimization}

\begin{algorithm}
    \caption{Optimization Algorithm of the DPE}
    \label{algo::optimization}
    \begin{algorithmic}[1]
        \Require A co-occurrence network between strings $N_{occur}$, a set of seed synonym pairs $S_{seed}$, a set of training patterns $S_{pat}$.
        \Ensure The string embeddings $\mathbf{x}$, parameters of the distributional score function $\mathbf{W_D}$, parameters of the pattern classifier $\mathbf{W_P}$.
        \While{$iter \leq I $}
            \State $\boxdot$ \textbf{\emph{Optimize $L_{C}$}}
            \State Sample a string pair $(u,v)$ from $N_{occur}$.
            \State Randomly sample $N$ negative string pairs $\{(u,v_n)\}_{n=1}^N$.
            \State Update $\mathbf{x},\mathbf{c}$ w.r.t. $L_{C}$.
            
            \State $\boxdot$ \textbf{\emph{Optimize $L_{S}$}}
            \State Sample a string pair $(u,v)$ from $S_{seed}$.
            \State Randomly sample a negative string pair $(u,v_n)$
            \State Update $\mathbf{x}$ and $\mathbf{W_D}$ w.r.t. $L_{S}$.
            
            \State $\boxdot$ \textbf{\emph{Optimize $O_{P}$}}
            \State Sample a pattern from $S_{pat}$.
            \State Update $\mathbf{x}$ and $\mathbf{W_P}$ w.r.t. $O_{P}$.
            
        \EndWhile
            
    \end{algorithmic}
\end{algorithm}

\smallskip
\noindent \textsf{\textbf{Synonym Inference.}}
To infer the synonyms of a query entity, our framework leverages both the distributional module and the pattern module.

Formally, given a query entity $e$, suppose its name strings collected from knowledge bases is $S_{syn}(e)$. Then for each candidate string $u$, we measure the possibility that $u$ is a synonym of $e$ using the following score function:
\begin{equation}
	\label{eqn::score}
	Score(e,u)=\sum_{s \in S_{syn}(e)} \{ Score_D(s,u) + \lambda Score_P(s,u) \}.
\end{equation}
$Score_D$ (Eqn.~\ref{eqn::score_d}) and $Score_P$ (Eqn.~\ref{eqn::score_p}) are used to measure how likely two target strings are synonymous, which are learned from the distributional module and the pattern module respectively. $\lambda$ is a parameter controlling the relative weights of the two parts.
The definition of the score function is quite intuitive. For each candidate string, we will compare it with all existing name strings of the query entity, and these existing name strings will vote to decide whether the candidate string is a synonym of the query entity.

However, the above method is not scalable. The reason is that the computational cost of the pattern score $Score_P$ is very high, as we need to collect and analyze all the sentences mentioning both the target strings. When the number of candidate strings is very large, calculating the pattern scores for all candidate strings can be very time-consuming. To solve the problem, as the distributional score $Score_D$ between two target strings is easy to calculate, a more efficient solution could be first utilizing the distributional score $Score_D$ to construct a set of high potential candidates, and then using the integrated score $Score$ to find the synonyms from those high potential candidates. 

Therefore, for each query entity $e$, we first rank each candidate string according to their distributional scores $Score_D$, and extract the top ranked candidate strings as the high potential candidates. After that, we re-rank the high potential candidates with the integrated score $Score$, and treat the top ranked candidate strings as the discovered synonym of entity $e$. With such two-step strategy, we are able to discover synonyms both precisely and efficiently.

\section{Experiment}

\subsection{Experiment Setup}
\subsubsection{\textbf{Datasets}}
Three datasets are constructed in our experiments.
(1) \textbf{Wiki + Freebase}: We treat the first 100K articles in the Wikipedia~\footnote{~\url{https://www.wikipedia.org/}} dataset as the text data, and the Freebase~\footnote{~\url{https://developers.google.com/freebase/}}~\cite{freebase:datadumps} as the knowledge base. 
(2) \textbf{PubMed + UMLS}: We collect around 1.5M paper abstracts from the PubMed dataset~\footnote{~\url{https://www.ncbi.nlm.nih.gov/pubmed}}, and use the UMLS~\footnote{~\url{https://www.nlm.nih.gov/research/umls/}} dataset as our knowledge base.
(3) \textbf{NYT + Freebase}: We randomly sample 118664 documents from 2013 New York Times news articles, and we select the Freebase as the knowledge base.
For each dataset, we adopt the Stanford CoreNLP package~\cite{manning-EtAl:2014:P14-5}\footnote{~\url{http://stanfordnlp.github.io/CoreNLP/}} to do tokenization, part-of-speech tagging and dependency parsing. We filtered out strings that appear less than 10 times. The window size is set as 5 when constructing the co-occurrence network between strings.
The statistics of the datasets are summarized in Table~\ref{tab::dataset}.

\begin{table}[!htb]
\caption{Statistics of the Datasets.}
	\label{tab::dataset}
	\centering
	\scalebox{1}{
        \begin{tabular}{c c c c}
        \hline
        \textbf{Dataset} & \textbf{Wiki} & \textbf{PubMed} & \textbf{NYT} \\
        \hline
        $\#$Documents & 100,000 & 1,554,433 & 118,664  \\ 
        $\#$Sentences & 6,839,331 & 15,051,203 & 3,002,123  \\ 
        $\#$Strings in Vocab & 277,635 & 357,985 & 115,680  \\ 
        $\#$Training Entities & 4,047 & 9,298 & 1,219  \\ 
        $\#$Test Entities (Warm) & 256 & 250 & 79  \\ 
        $\#$Test Entities (Cold) & 175 & 150 & 72  \\ 
        \hline
        \end{tabular}
    }
\end{table}

\begin{table*} [!htb]
	\caption{Quantitative results on the warm-start setting.}
	\label{tab::results-warm}
	\begin{center}
		\scalebox{0.8}{
		\begin{tabular}{c|c c c c c c|c c c c c c|c c c c c c}\hline
		\multirow{2}{*}{\textbf{Algorithm}}	& \multicolumn{6}{c|}{\textbf{Wiki + Freebase}} & \multicolumn{6}{c|}{\textbf{PubMed + UMLS}} & \multicolumn{6}{c}{\textbf{NYT + Freebase}}\\ 
		& \textbf{P@1}& \textbf{R@1} &  \textbf{F1@1} & \textbf{P@5}& \textbf{R@5} &  \textbf{F1@5} & \textbf{P@1}& \textbf{R@1} &  \textbf{F1@1} & \textbf{P@5}& \textbf{R@5} &  \textbf{F1@5} & \textbf{P@1}& \textbf{R@1} &  \textbf{F1@1}& \textbf{P@5}& \textbf{R@5} &  \textbf{F1@5} \\ \hline 
	Patty& 0.102 & 0.075 & 0.086 & 0.049 & 0.167 & 0.076 & 0.352 & 0.107 & 0.164 & 0.164 & 0.248 & 0.197 & 0.101 & 0.081 & 0.090 & 0.038 & 0.141 & 0.060 \\  
	SVM& 0.508 & 0.374 & 0.431 & 0.273 & 0.638 & 0.382 & 0.696 & 0.211 & 0.324 & 0.349 & 0.515 & 0.416 & 0.481 & 0.384 & 0.427 & 0.248 & 0.616 & 0.354  \\ 
	word2vec& 0.387 & 0.284 & 0.328 & 0.247 & 0.621 & 0.353 & 0.784 & 0.238 & 0.365 & 0.464 & 0.659 & 0.545 & 0.367 & 0.293 & 0.326 & 0.216 & 0.596 & 0.317  \\ 
	GloVe& 0.254 & 0.187 & 0.215 & 0.104 & 0.316 & 0.156 & 0.536 & 0.163 & 0.250 & 0.279 & 0.417 & 0.334 & 0.203 & 0.162 & 0.180 & 0.084 & 0.283 & 0.130  \\ 
	PTE& 0.445 & 0.328 & 0.378 & 0.252 & 0.612 & 0.357 & 0.800 & 0.243 & 0.373 & 0.476 & 0.674 & 0.558 & 0.456 & 0.364 & 0.405 & 0.233 & 0.606 & 0.337  \\ 
	RKPM& 0.500 & 0.368 & 0.424 & 0.302 & 0.681 & 0.418 & 0.804 & 0.244 & 0.374 & 0.480 & 0.677 & 0.562 & 0.506 & 0.404 & 0.449 & 0.302 & 0.707 & 0.423  \\ \hline	
    DPE-NoP & 0.641 & 0.471 & 0.543 & 0.414 & 0.790 & 0.543 & 0.816 & 0.247 & 0.379 & 0.532 & 0.735 & 0.617 & 0.532 & 0.424 & 0.472 & 0.305 & 0.687 & 0.422\\ 
    DPE-TwoStep & 0.684 & 0.503 & 0.580 & 0.417 & 0.782 & 0.544 & 0.836 & 0.254 & 0.390 & 0.538 & 0.744 & 0.624 & 0.557 & 0.444 & 0.494 & 0.344 & 0.768 & 0.475\\
    DPE & \textbf{0.727} & \textbf{0.534} & \textbf{0.616} & \textbf{0.465} & \textbf{0.816} & \textbf{0.592} & \textbf{0.872} & \textbf{0.265} & \textbf{0.406} & \textbf{0.549} & \textbf{0.755} & \textbf{0.636} & \textbf{0.570} & \textbf{0.455} & \textbf{0.506} & \textbf{0.366} & \textbf{0.788} & \textbf{0.500}\\ \hline	
	\end{tabular}
	}
	\end{center}
\end{table*}

\begin{table*} [!htb]
	\caption{Quantitative results on the cold-start setting.}
	\label{tab::results-cold}
	\begin{center}
		\scalebox{0.8}{
		\begin{tabular}{c|c c c c c c|c c c c c c|c c c c c c}\hline
		\multirow{2}{*}{\textbf{Algorithm}}	& \multicolumn{6}{c|}{\textbf{Wiki + Freebase}} & \multicolumn{6}{c|}{\textbf{PubMed + UMLS}} & \multicolumn{6}{c}{\textbf{NYT + Freebase}}\\
		& \textbf{P@1}& \textbf{R@1} &  \textbf{F1@1} & \textbf{P@5}& \textbf{R@5} &  \textbf{F1@5} & \textbf{P@1}& \textbf{R@1} &  \textbf{F1@1} & \textbf{P@5}& \textbf{R@5} &  \textbf{F1@5} & \textbf{P@1}& \textbf{R@1} &  \textbf{F1@1}& \textbf{P@5}& \textbf{R@5} &  \textbf{F1@5} \\ \hline 
	Patty& 0.131 & 0.056 & 0.078 & 0.065 & 0.136 & 0.088 & 0.413 & 0.064 & 0.111 & 0.191 & 0.148 & 0.167 & 0.125 & 0.054 & 0.075 & 0.062 & 0.132 & 0.084  \\  
	SVM& 0.371 & 0.158 & 0.222 & 0.150 & 0.311 & 0.202 & 0.707 & 0.110 & 0.193 & 0.381 & 0.297 & 0.334 & 0.347 & 0.150 & 0.209 & 0.165 & 0.347 & 0.224  \\ 
	word2vec& 0.411 & 0.175 & 0.245 & 0.196 & 0.401 & 0.263 & 0.627 & 0.098 & 0.170 & 0.408 & 0.318 & 0.357 & 0.361 & 0.156 & 0.218 & 0.151 & 0.317 & 0.205  \\ 
	GloVe& 0.251 & 0.107 & 0.150 & 0.105 & 0.221 & 0.142 & 0.480 & 0.075 & 0.130 & 0.264 & 0.206 & 0.231 & 0.181 & 0.078 & 0.109 & 0.084 & 0.180 & 0.115  \\ 
	PTE& 0.474 & 0.202 & 0.283 & 0.227 & 0.457 & 0.303 & 0.647 & 0.101 & 0.175 & 0.389 & 0.303 & 0.341 & 0.403 & 0.174 & 0.243 & 0.166 & 0.347 & 0.225  \\ 
	RKPM& 0.480 & 0.204 & 0.286 & 0.227 & 0.455 & 0.303 & 0.700 & 0.109 & 0.189 & 0.447 & 0.348 & 0.391 & 0.403 & 0.186 & 0.255 & 0.170 & 0.353 & 0.229  \\ \hline	
    DPE-NoP & 0.491 & 0.209 & 0.293 & 0.246 & 0.491 & 0.328 & 0.700 & 0.109 & 0.189 & 0.456 & 0.355 & 0.399 & 0.417 & 0.180 & 0.251 & 0.180 & 0.371 & 0.242\\ 
    DPE-TwoStep & 0.537 & 0.229 & 0.321 & 0.269 & 0.528 & 0.356 & 0.720 & 0.112 & 0.194 & 0.477 & 0.372 & 0.418 & 0.431 & 0.186 & 0.260 & 0.183 & 0.376 & 0.246\\
    DPE & \textbf{0.646} & \textbf{0.275} & \textbf{0.386} & \textbf{0.302} & \textbf{0.574} & \textbf{0.396} & \textbf{0.753} & \textbf{0.117} & \textbf{0.203} & \textbf{0.500} & \textbf{0.389} & \textbf{0.438} & \textbf{0.486} & \textbf{0.201} & \textbf{0.284} & \textbf{0.207} & \textbf{0.400} & \textbf{0.273}\\ \hline	
	\end{tabular}
	}
	\end{center}
\end{table*}

\begin{figure*}[htb!]
    \vspace{-0.5cm}
	\centering
	\subfigure[Precision (Warm Start)]{
		\includegraphics[width=0.2\textwidth]{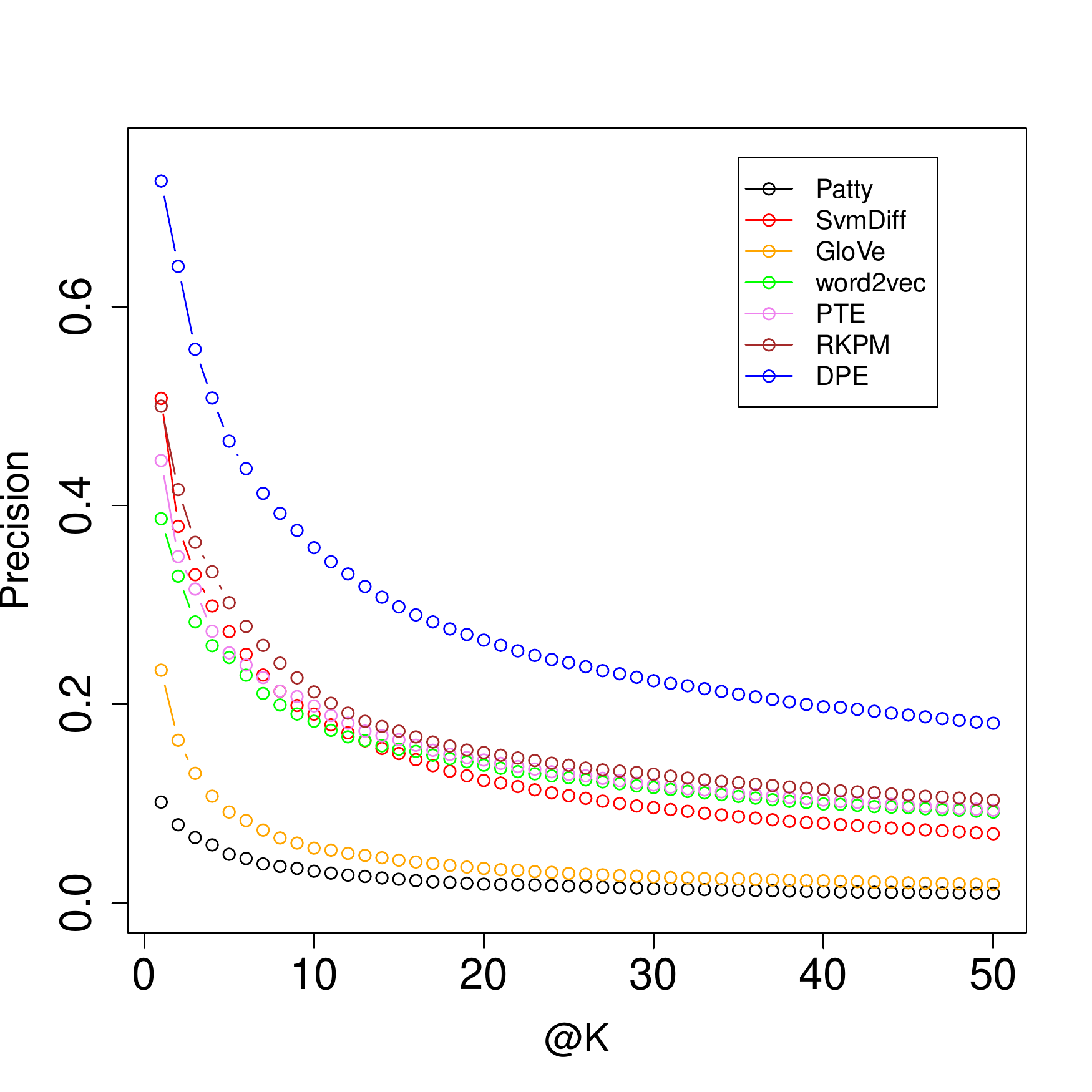}	
	}
	\hspace{0.5cm}
	\subfigure[Recall (Warm Start)]{
		\includegraphics[width=0.2\textwidth]{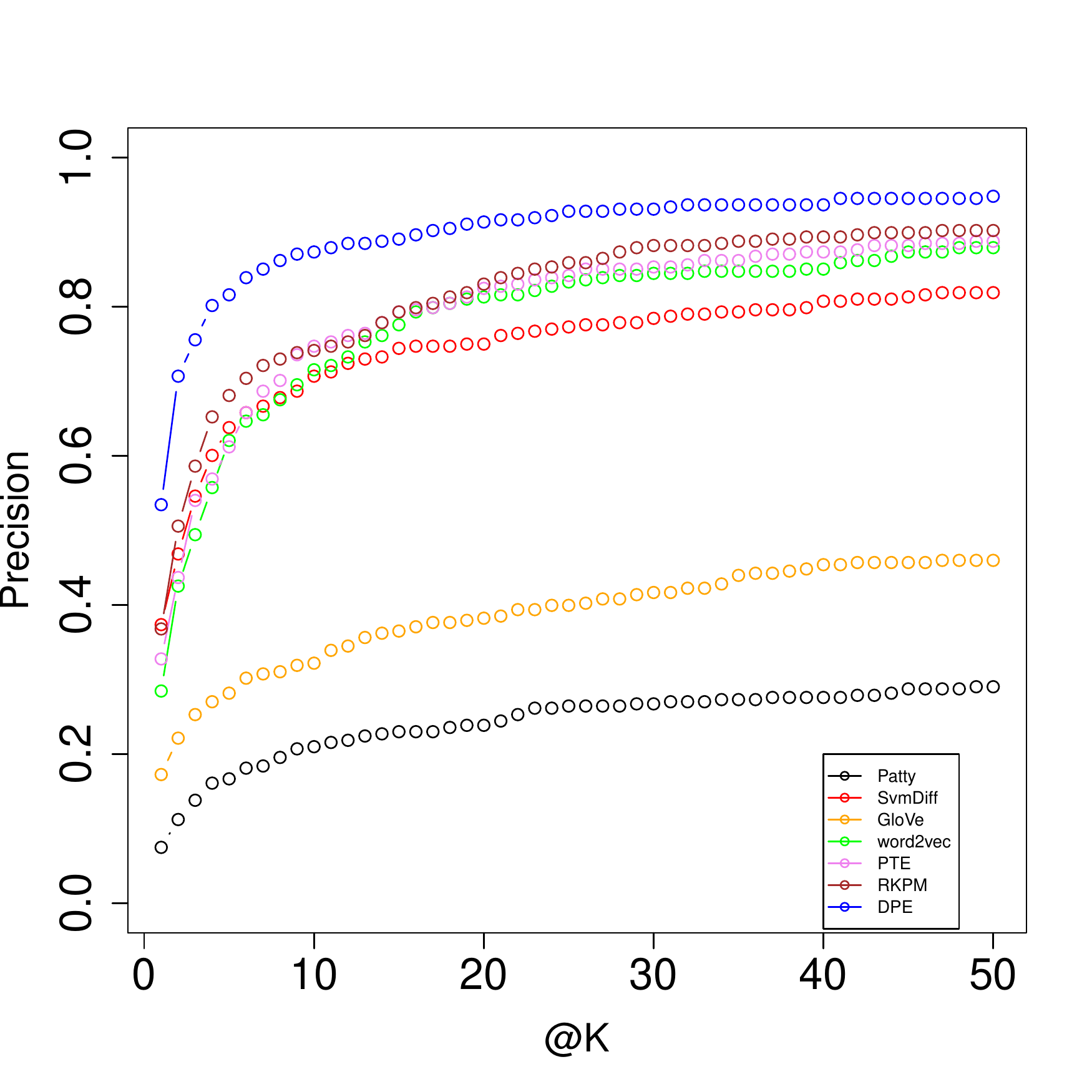}
	}
	\hspace{0.5cm}
	\subfigure[Precision (Cold Start)]{
		\includegraphics[width=0.2\textwidth]{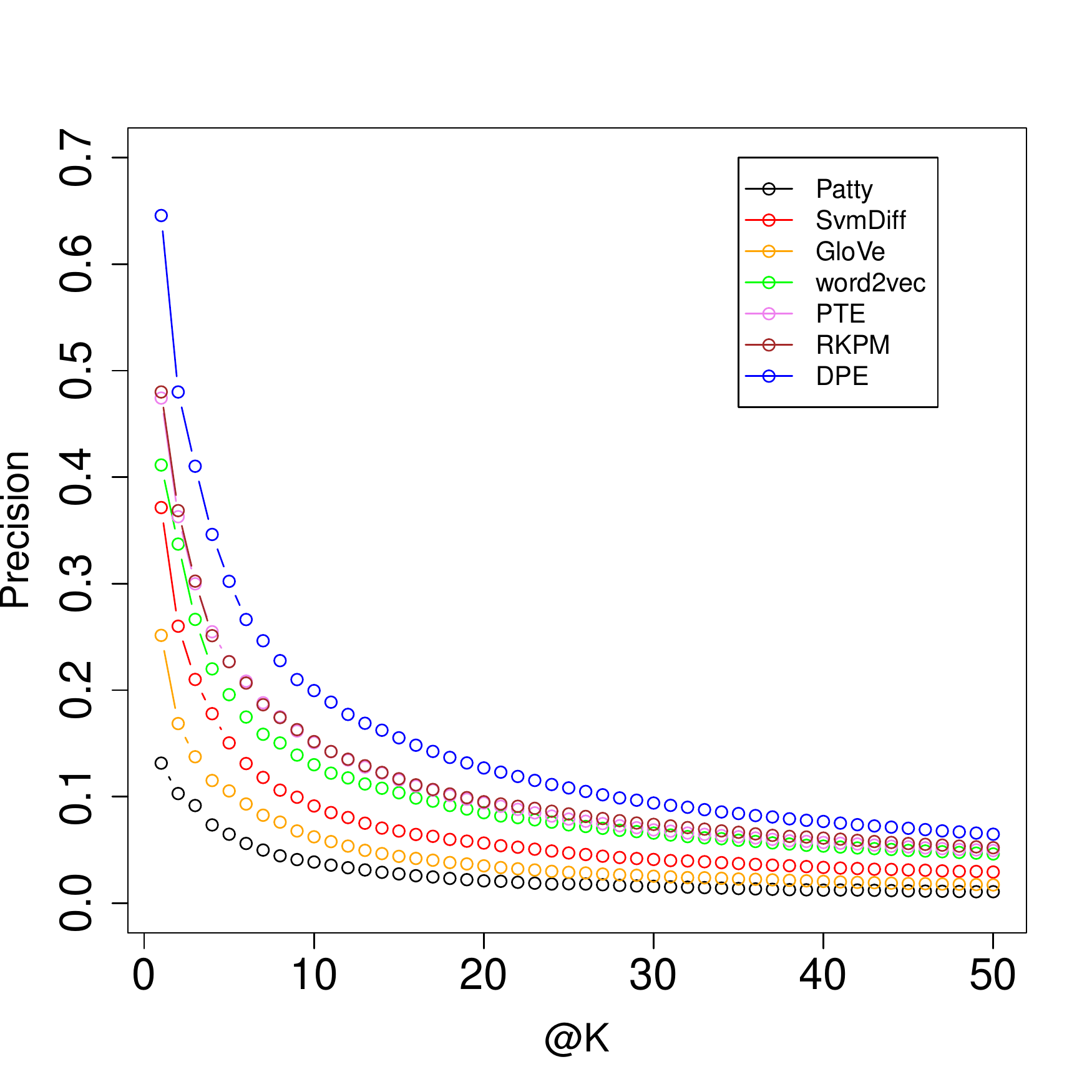}	
	}
	\hspace{0.5cm}
	\subfigure[Recall (Cold Start)]{
		\includegraphics[width=0.2\textwidth]{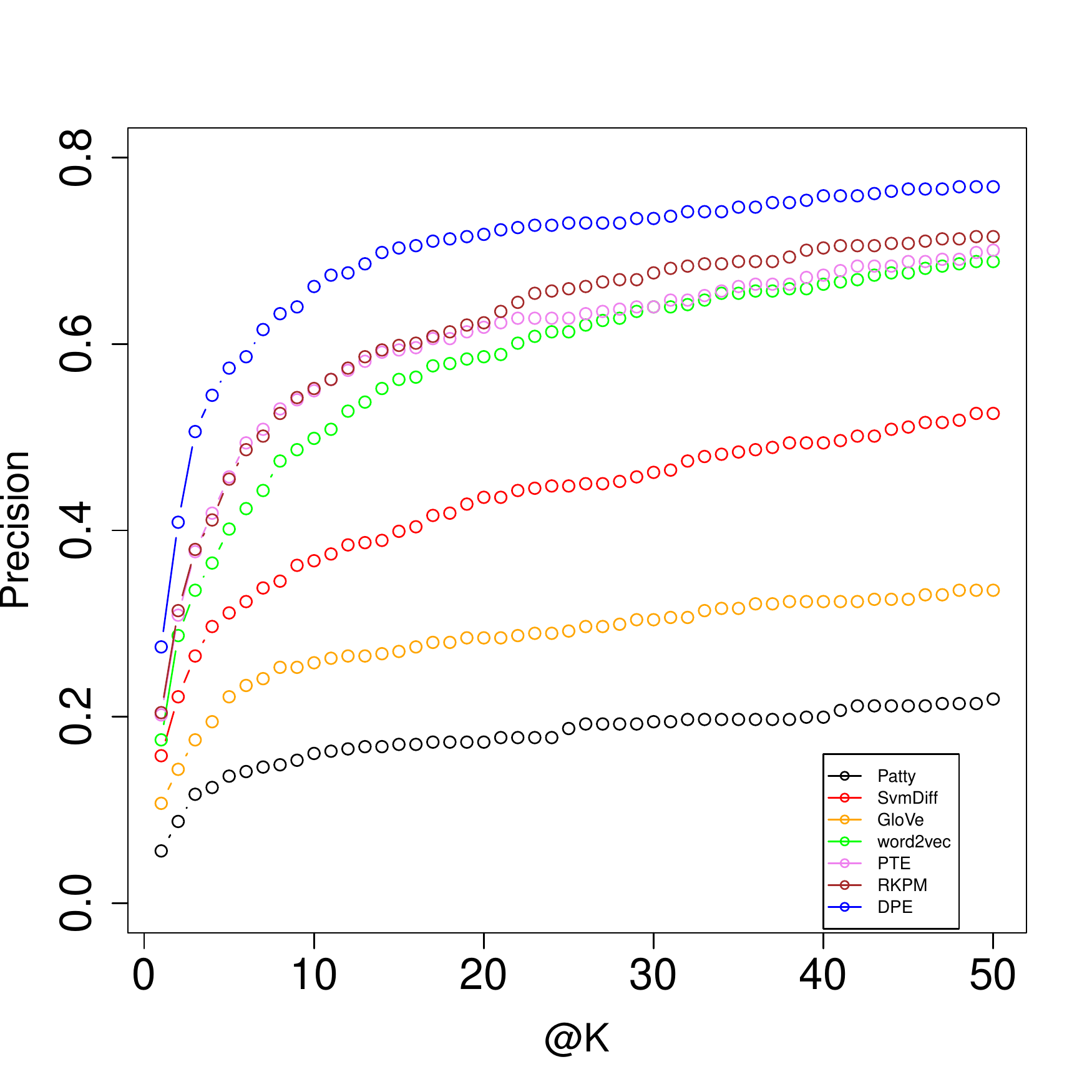}
	}
	\caption{Precision and Recall at different positions on the Wiki dataset.}
	\label{fig::mcts}
	\vspace{-0.3cm}
\end{figure*}

\subsubsection{\textbf{Performance Evaluation}}

For each dataset, we randomly sample some linked entities as the training entities, and all their synonyms are used as seeds by the compared approaches. We also randomly sample a few linked entities as test entities, which are used for evaluation.

Two settings are considered in our experiments, \textit{i.e.}, the warm-start setting and the cold-start setting. In the warm-start setting, for each test entity, we assume that 50\% of its synonyms are already given, and we aim to use them to infer the rest 50\%. In the cold-start setting, we are only given the original name of each test entity, and our goal is to infer all its synonyms in knowledge bases.

During evaluation, we treat all unlinkable strings (\textit{i.e.}, words or phrases that are not linked to any entities in the knowledge base) as the candidate strings. In both settings, we add the ground-truth synonyms of each test entity into the set of candidate strings, and we aim to rank the ground-truth synonyms at the top positions among all candidate strings. For the evaluation metrics, we report the Precision at Position $K$ (P@$K$), Recall at Position $K$ (R@$K$) and F1 score at Position $K$ (F1@$K$). 

\subsubsection{\textbf{Compared algorithms}}
We select the following algorithms to compare.
(1) \textbf{Patty}~\cite{nakashole2012patty}: a pattern based approach for relation extraction, which can be applied to our problem by treating the collected synonym seeds as training instances.
(2) \textbf{SVM}~\cite{weeds2014learning}: a distributional based approach, which uses the bag-of-words features and learns an SVM classifier for synonym discovery. 
(3) \textbf{word2vec}~\cite{mikolov2013distributed}: a word embedding approach. We use the learned string embedding as features and train a score function in Eqn.~\ref{eqn::score_d} for synonym discovery.
(4) \textbf{GloVe}~\cite{pennington2014glove}: another word embedding approach. Similar to word2vec, we use the learned string embedding as features and train a score function for synonym discovery.
(5) \textbf{PTE}~\cite{tang2015pte}: a text embedding approach, which is able to exploit both the text data and the entity types provided in knowledge bases to learn string embeddings. After embedding learning, we apply the score function in Eqn.~\ref{eqn::score_d} for synonym discovery.
(6) \textbf{RKPM}~\cite{wang2015solving}: a knowledge powered string embedding approach, which utilizes both the raw text and the synonym seeds for synonym discovery.
(7) \textbf{DPE}: our proposed embedding framework, which integrates both the distributional features and local patterns for synonym discovery.
(8) \textbf{DPE-NoP}: a variant of our framework, which only deploys the distributional module ($O_D$).
(9) \textbf{DPE-TwoStep}: a variant of our framework, which first trains the distributional module ($O_D$) and then the pattern module ($O_P$), without jointly optimizing them.

\subsubsection{\textbf{Parameter Settings}}
For all embedding based approaches, we set the embedding dimension as 100.
For DPE and its variants, we set the learning rate as 0.01 and the number of negative samples $N$ when optimizing the co-occurrence network $L_D$ is set as 5.
When collecting the syntactic features in the pattern module, we set the n-gram length $N$ as 3.
The parameter $\lambda$, which controls the weights of the two modules during synonym discovery, is set as 0.1 by default. We set the number of iterations as 10 billions. During synonym inference, we first adopt the distributional module to extract top 100 ranked strings as the high potential candidates, then we use both modules to re-rank them.
For word2vec, PTE, the number of negative examples is also set as 5, and the initial learning rate is set as 0.025, as suggested by~\cite{tang2015line,tang2015pte,mikolov2013distributed}. The number of iterations is set as 20 for word2vec, and for PTE we sample 10 billion edges to ensure convergence.
For GloVe, we use the default parameter settings as used in~\cite{pennington2014glove}.
For RKPM, we set the learning rate as 0.01, and the iteration is set as 10 billion to ensure convergence.

\subsection{Experiments and Performance Study}

\smallskip
\noindent \textsf{\textbf{1. Comparing DPE with other baseline approaches.}}
Table~\ref{tab::results-warm}, Table~\ref{tab::results-cold} and Figure~\ref{fig::mcts} present the results on the warm-start and cold-start settings. In both settings, we see that the pattern based approach Patty does not perform well, and our proposed approach DPE significantly outperforms Patty. This is because most synonymous strings will never co-appear in any sentences, leading to the low recall of Patty. Also, many patterns discovered by Patty are not so reliable, which may harm the precision of the discovered synonyms.
DPE addresses this problem by incorporating the distributional information, which can effectively complement and regulate the pattern information, leading to higher recall and precision. 

Comparing DPE with the distributional based approaches (word2vec, GloVe, PTE, RKPM), DPE still significantly outperforms them. The performance gains mainly come from: (1) we exploit the co-occurrence observation~\ref{obs::co-occur} during training, which enables us to better capture the semantic meanings of different strings; (2) we incorporate the pattern information to improve the performances.

\smallskip
\noindent \textsf{\textbf{2. Comparing DPE with its variants.}}
To better understand why DPE achieves better results, we also compare DPE with several variants.
From Table~\ref{tab::results-warm} and Table~\ref{tab::results-cold}, we see that in most cases, the distributional module of our approach (DPE-NoP) can already outperform the best baseline approach RKPM. This is because we utilize the co-occurrence observation~\ref{obs::co-occur} in our distributional module, which helps us capture the semantic meanings of strings more effectively. 
By separately training the pattern module after the distributional module, and using both modules for synonym discovery (DPE-TwoStep), we see that the results are further improved, which demonstrates that the two modules can indeed mutually complement each other for synonym discovery.
If we jointly train both modules (DPE), we obtain even better results, which shows that our proposed joint optimization framework can benefit the training process and therefore helps achieve better results.

\begin{figure}[htb!]
	\centering
	\subfigure[Wiki (warm-start)]{
		\label{fig::lambda-w}
		\includegraphics[width=0.2\textwidth]{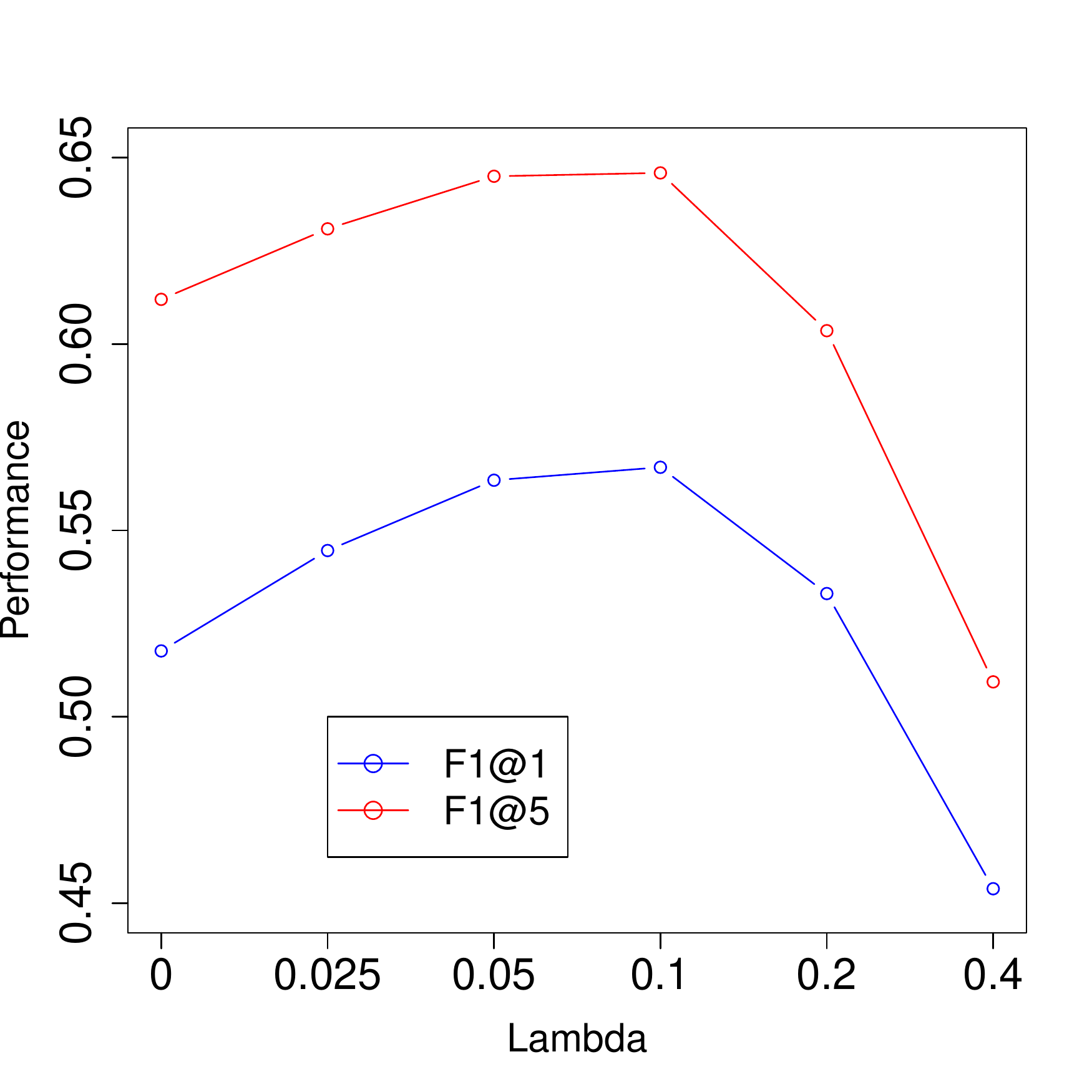}	
	}
	\subfigure[Wiki (cold-start)]{
		\label{fig::lambda-c}
		\includegraphics[width=0.2\textwidth]{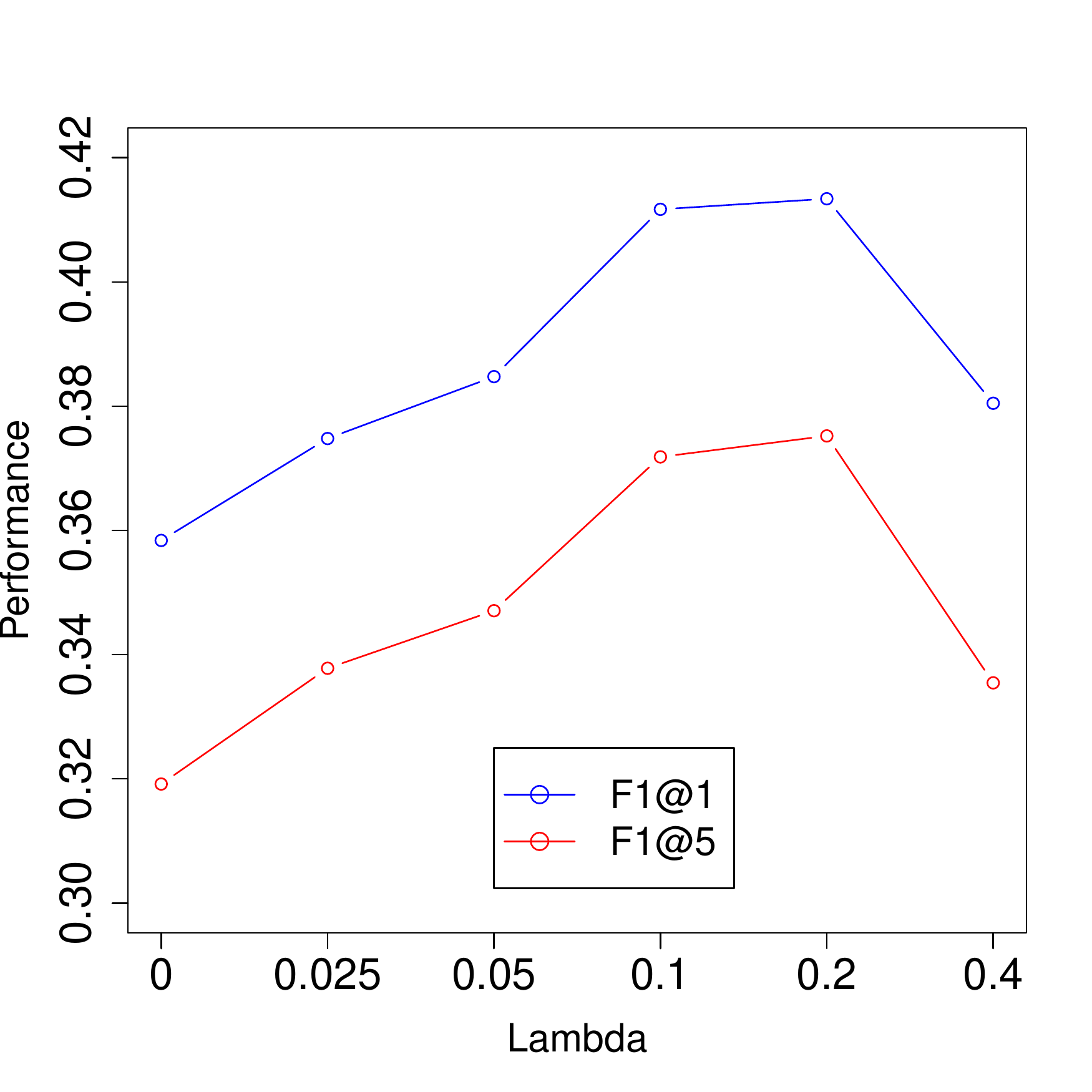}	
	}
	\caption{Performances w.r.t. $\lambda$. A small $\lambda$ emphasizes the distributional module. A large $\lambda$ emphasizes the pattern module. Either module cannot discover synonyms effectively.}
	\label{fig::lambda}
\end{figure}

\begin{figure}[htb!]
	\centering
	\vspace{-0.5cm}
	\subfigure[Percentage of Training Entities]{
		\label{fig::label}
		\includegraphics[width=0.2\textwidth]{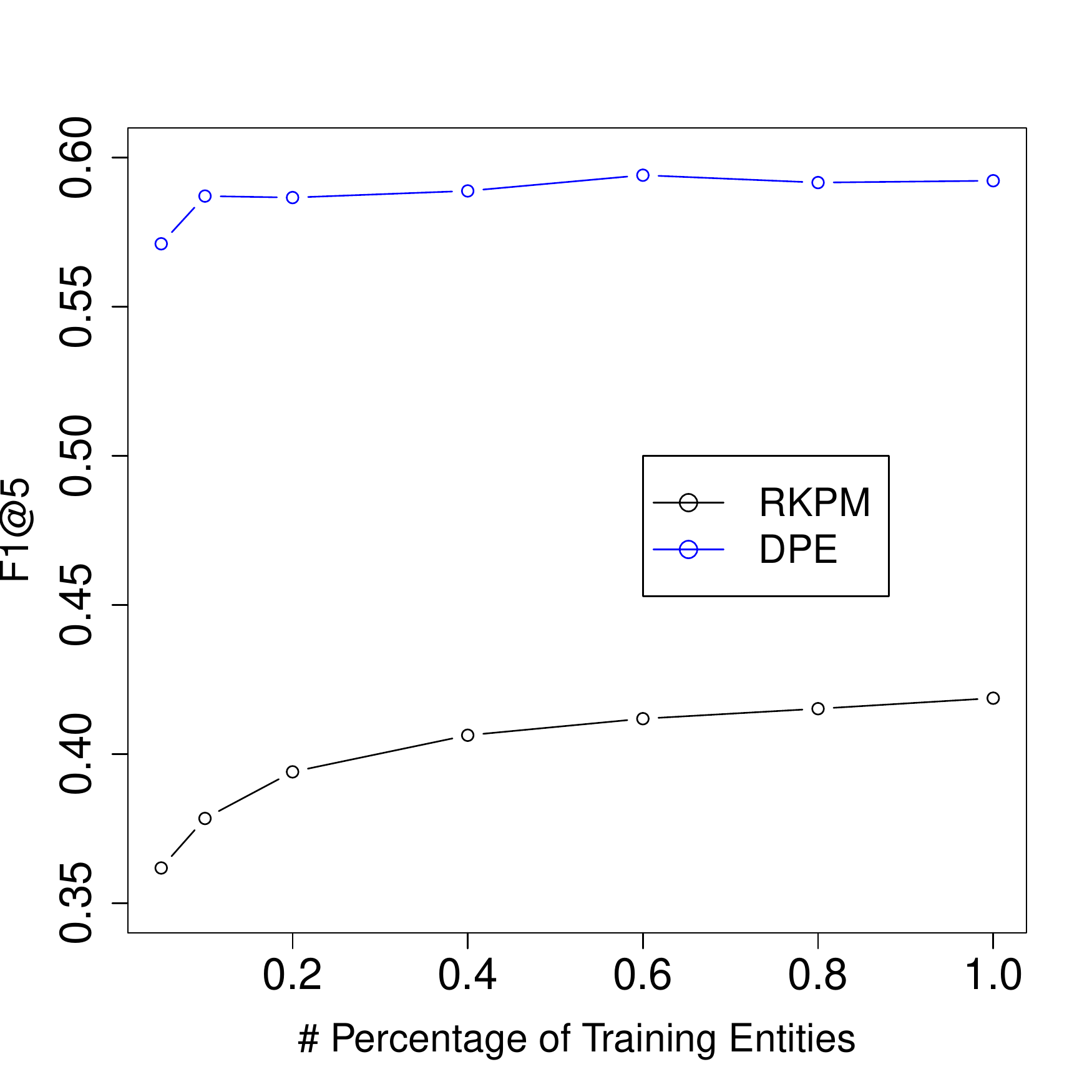}	
	}
	\subfigure[\#Synonyms Used in Inference]{
		\label{fig::sparsity}
		\includegraphics[width=0.2\textwidth]{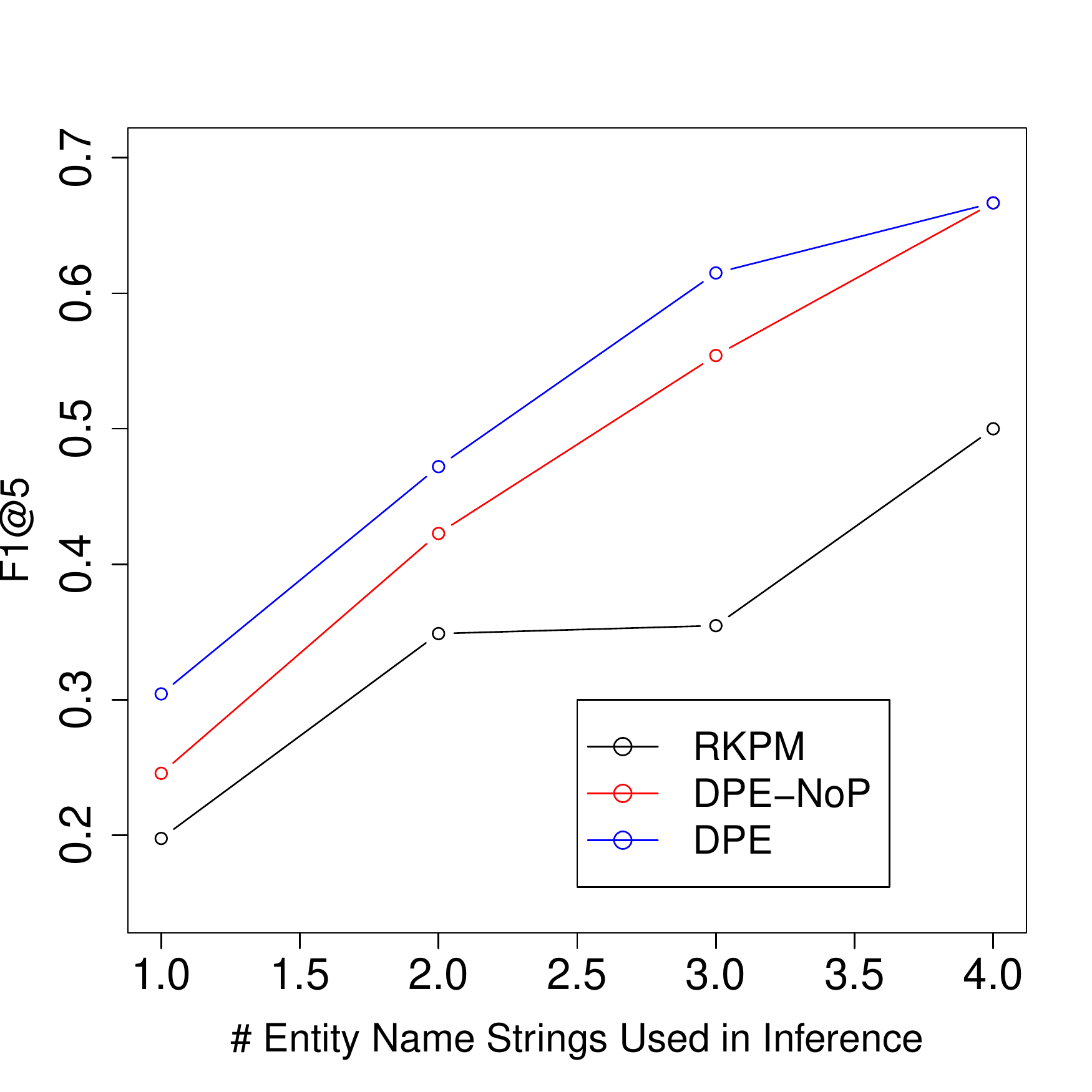}	
	}
	\caption{Performance change of DPE (a) under different percentage of training entities; and (b) with respect to the number of entity name strings used in inference.}
	\label{fig::performance}
\end{figure}

\smallskip
\noindent \textsf{\textbf{3. Performances w.r.t. the weights of the modules.}}
During synonym discovery, DPE will consider the scores from both the distributional module and the pattern module, and the parameter $\lambda$ controls the relative weight. Next, we study how DPE behaves under different $\lambda$.
The results on the Wiki dataset are presented in Figure~\ref{fig::lambda}. We see that when $\lambda$ is either small or large, the performance is not so good. This is because a small $\lambda$ will emphasize only the distributional module, while a large $\lambda$ will assign too much weight to the pattern module. Therefore, either the distributional module or the pattern module cannot discover synonyms effectively, and we must integrate them during synonym discovery.

\smallskip
\noindent \textsf{\textbf{4. Performances w.r.t. the percentage of the training entities.}} 
During training, DPE will use the synonyms of the training entities as seeds to guide the training.
To understand how the training entities will affect the results, we report the performances of DPE under different percentages of training entities. Figure~\ref{fig::label} presents the results on the Wiki dataset under the warm-start setting. We see that compared with RKPM, DPE needs fewer labeled data to converge. This is because the two modules in our framework can mutually complement each other, and therefore reduce the demand of the training entities.

\smallskip
\noindent \textsf{\textbf{5. Performances w.r.t. the number of entity name strings used in inference.}} 
Our framework aims to discover synonyms at the entity level. Specifically, for each query entity, we use its existing name strings to disambiguate the meaning for each other, and let them vote to discover the missing synonyms. In this section, we study how the number of name strings in inference will affect the results. We sample a number of test entities from the Wiki dataset, and utilize 1$\sim$4 existing name strings of each entity to do inference. Figure~\ref{fig::sparsity} presents the results.
We see that DPE consistently outperforms RKPM. Besides, DPE also outperforms its variant DPE-NoP, especially when the number of name strings used in inference is small. The reason may be that the pattern module of DPE can effectively complement the distributional module when only few entity name strings are available during inference.

\begin{table}[!htb]
\caption{Example outputs on the Wiki dataset. Strings with red colors are the true synonyms.}
	\label{tab::case}
	\centering
	\scalebox{0.8}{
        \begin{tabular}{c|c c|c c}
        \hline
        \textbf{Entity} & \multicolumn{2}{c|}{\textbf{US dollar}} & \multicolumn{2}{c}{\textbf{World War II}}  \\ \hline
        \textbf{Method} & \textbf{DPE-NoP} & \textbf{DPE} & \textbf{DPE-NoP} & \textbf{DPE} \\
        \hline
        \multirow{5}{*}{\textbf{Output}} & {\color{red}{US Dollars}} & {\color{red}{U.S. dollar}} & {\color{red}{Second World War}} & {\color{red}{Second World War}} \\ 
         & {\color{red}{U.S. dollars}} & {\color{red}{US dollars}} & {\color{red}{World War Two}} & {\color{red}{World War Two}} \\ 
         & Euros & {\color{red}{U.S. dollars}} & World War One & {\color{red}{WW II}}  \\ 
         & {\color{red}{U.S. dollar}} & {\color{red}{U.S. \$}} & WW I & world war \\ 
         & RMB & Euros & world wars & world wars  \\ 
        \hline
        
        \end{tabular}
        \vspace{-0.3cm}
    }
\end{table}

\begin{table}[!htb]
\caption{Top ranked patterns expressing the synonym relation. Strings with red colors are the target strings.}
	\label{tab::pattern}
	\centering
	\scalebox{0.9}{
        \begin{tabular}{c c}
        \hline
        \textbf{Pattern} & \textbf{Corresponding Sentence}  \\
        \hline
        (-,NN,nsubj) (-lrb-,JJ,amod) & ... {\color{red}{Olympia}} ( commonly known as   \\ 
       (known,VBN,acl) (-,NN,nmod) & {\color{red}{L'Olympia}} ) is a music hall ... \\
       \hline
        \multirow{2}{*}{(-,NN,dobj) (-,NN,appos)} & ... , many hippies used {\color{red}{cannabis}}  \\ 
         & ( {\color{red}{marijuana}} ) , considering it ... \\
       \hline
        (-,NNP,nsubj) (known,VBN,acl) & ... {\color{red}{BSE}} , commonly known as ''   \\ 
        (-,NN,nmod) & {\color{red}{mad cow disease}} '' , is a ... \\
        \hline
        
        \end{tabular}
    }
\end{table}

\subsection{Case Studies}

\smallskip
\noindent \textsf{\textbf{1. Example output.}}
Next, we present some example outputs of DPE-NoP and DPE on the Wiki dataset. The results are shown in Figure~\ref{tab::case}. From the learned synonym list, we have filtered out all existing synonyms in knowledge bases, and the red strings are the new synonyms discovered by our framework. We see that our framework finds many new synonyms which have not been included in knowledge bases. Besides, by introducing the pattern module, we see that some false synonyms (RMB and WW I) obtained by DPE-NoP will be filtered out by DPE, which demonstrates that combing the distributional features and the local patterns can indeed improve the performances.

\smallskip
\noindent \textsf{\textbf{2. Top ranked positive patterns.}}
To exploit the local patterns in our framework, our pattern module learns a pattern classifier to predict whether a pattern expresses the synonym relation between the target strings. To test whether the learned classifier can precisely discover some positive patterns for synonym discovery, we show some top-ranked positive patterns learned by the classifier and also the corresponding sentences. Table~\ref{tab::pattern} presents the results, in which the red strings are the target strings. We see that all the three patterns indeed express the synonym relations between the target strings, which proves that our learned pattern classifier can effectively find some positive patterns and therefore benefit the synonym discovery.

\section{Related Work}
\label{sec::related}

\noindent \textsf{\textbf{Synonym Discovery.}}
Various approaches have been proposed to discover synonyms from different kinds of information.
Most of them exploit structured knowledge such as query logs~\cite{ren2015synonym,chaudhuri2009exploiting,wei2009context} for synonym discovery. Different from them, we aim to discover synonyms from raw text corpora, which is more challenging.

There are also some methods trying to discover string relations (\textit{e.g.}, synonym relation, antonym relation, hypernym relation) from raw texts, including some distributional based approaches and pattern based approaches. Both approaches can be applied to our setting.
Given some training seeds, the distributional based approaches~\cite{wang2015solving,weeds2014learning,roller2014inclusive,turney2001mining,turney2001mining,lin2003identifying,pantel2009web} discover synonyms by representing strings with their distributional features, and learning a classifier to predict the relation between strings.
Different from them, the pattern based approaches~\cite{yahya2014renoun,nakashole2012patty,snow2004learning,sun2010semi,qian2009semi,hearst1992automatic} consider the sentences mentioning a pair of synonymous strings, and learn some textual patterns from these sentences, which are further used to discover more synonyms.
Our proposed approach naturally integrates the two types of approaches, which enjoys both merits of them.

\noindent \textsf{\textbf{Text Embedding.}}
Our work is also related to text embedding techniques, which learn low-dimensional vector representations for strings from raw texts. 
The learned embedding capture some semantic correlations between strings, which can be used as features for synonym extraction.
Most text embedding approaches~\cite{mikolov2013distributed,pennington2014glove,tang2015line} only exploit the text data, which cannot exploit information from knowledge bases to guide the embedding learning.
There are also some studies trying to incorporate knowledge bases to improve the embedding learning. ~\cite{tang2015pte,ren2016label} exploit entity types to enhance the learned embedding and~\cite{weston2013connecting,xu2014rc,wang2014knowledge,liu2015learning} exploit existing relation facts in knowledge bases as constraints to improve the performances.

Compared with these methods, our embedding approach can better preserve the semantic correlations of strings with the the co-occurrence observation~\ref{obs::co-occur}. Besides, both the distributional module and the pattern module of our approach will provide supervision for embedding learning, which brings stronger predictive abilities to the learned embeddings under the synonym discovery problem.

\section{Conclusions}
In this paper, we studied the problem of automatic synonym discovery with knowledge bases, aiming to discover missing synonyms for entities in knowledge bases. We proposed a framework called the DPE, which naturally integrates the distributional based approaches and the pattern based approaches. We did extensive experiments on three real-world datasets. Experimental results proved the effectiveness of our proposed framework.

\section*{Acknowledgments}
Research was sponsored in part by the U.S. Army Research Lab. under Cooperative Agreement No. W911NF-09-2-0053 (NSCTA), National Science Foundation IIS-1320617 and IIS 16-18481, and grant 1U54GM114838 awarded by NIGMS through funds provided by the trans-NIH Big Data to Knowledge (BD2K) initiative (www.bd2k.nih.gov). The views and conclusions contained in this document are those of the author(s) and should not be interpreted as representing the official policies of the U.S. Army Research Laboratory or the U.S. Government. The U.S. Government is authorized to reproduce and distribute reprints for Government purposes notwithstanding any copyright notation hereon.

\bibliographystyle{abbrv}
\bibliography{sigproc}
\end{document}